\documentclass[conference]{IEEEtran}
\IEEEoverridecommandlockouts
\usepackage{cite}
\usepackage{amsmath,amssymb,amsfonts}
\usepackage{algorithm}
\usepackage[noend]{algpseudocode}
\usepackage{graphicx}
\usepackage{textcomp}
\usepackage{xcolor}
\usepackage[bookmarks=false]{hyperref}       
\usepackage{url}            
\usepackage{float}

\ifCLASSOPTIONcompsoc
    \usepackage[caption=false,font=normalsize,labelfont=sf,textfont=sf]{subfig}
\else
    \usepackage[caption=false,font=footnotesize]{subfig}
\fi

\def\BibTeX{{\rm B\kern-.05em{\sc i\kern-.025em b}\kern-.08em
    T\kern-.1667em\lower.7ex\hbox{E}\kern-.125emX}}

\begin{document}

\title{A Conceptual Bio-Inspired Framework for the Evolution of Artificial General Intelligence}


\author{\IEEEauthorblockN{Sidney Pontes-Filho\IEEEauthorrefmark{1}\textsuperscript{,}\IEEEauthorrefmark{2} and Stefano Nichele\IEEEauthorrefmark{1}\textsuperscript{,}\IEEEauthorrefmark{3}}
\IEEEauthorblockA{\IEEEauthorrefmark{1}\textit{Department of Computer Science, Oslo Metropolitan University, Oslo, Norway}\\
\IEEEauthorrefmark{2}\textit{Department of Computer Science, Norwegian University of Science and Technology, Trondheim, Norway}\\
\IEEEauthorrefmark{3}\textit{Department of Holistic Systems, Simula Metropolitan, Oslo, Norway}\\
Email: sidneyp@oslomet.no, stenic@oslomet.no}}

\maketitle
\thispagestyle{plain}
\pagestyle{plain}

\begin{abstract}
In this work, a conceptual bio-inspired parallel and distributed learning framework for the emergence of general intelligence is proposed, where agents evolve through environmental rewards and learn throughout their lifetime without supervision, i.e., self-learning through embodiment. The chosen control mechanism for agents is a biologically plausible neuron model based on spiking neural networks. Network topologies become more complex through evolution, i.e., the topology is not fixed, while the synaptic weights of the networks cannot be inherited, i.e., newborn brains are not trained and have no innate knowledge of the environment. What is subject to the evolutionary process is the network topology, the type of neurons, and the type of learning. This process ensures that controllers that are passed through the generations have the intrinsic ability to learn and adapt during their lifetime in mutable environments. We envision that the described approach may lead to the emergence of the simplest form of artificial general intelligence.
\end{abstract}


\section{Introduction}
\label{sec:intro}

The brain is a truly remarkable computing machine that continuously adapts through sensory inputs. Rewards and penalties are encoded and learned throughout the evolution of organisms living in an environment (our world) that continuously provides unlabeled and mutable data. The supervision in the brain is a product of such evolutionary process. A real-world environment does not provide labeled data or predefined fitness functions for organisms and their brains as in supervised and reinforcement learning in Artificial Intelligence (AI) systems. However, organisms selected by natural and sexual selection \cite{arnold1984measurement} know or are able to learn which sensory inputs or input sequences may affect positively or negatively their survival and reproduction. One of the key components which ensures that a species will reproduce is the lifetime of the organisms. Pleasure, joy, and desire (or other positive inputs) may increase the lifetime of an organism and act as rewards. Pain, fear, and disgust may decrease the lifetime and act as penalties. All those feelings and emotions are results of the evolutionary pressure for increasing the life expectancy and succeeding in generating offspring \cite{Zador582643}. One example is the desire and disgust that arise for some smells. The desire may come from the smell of nutritive food which increases life expectancy, while the disgust may come from spoiled food which may cause food poisoning, and therefore causing a lifetime reduction. Evolution by natural selection made it possible for living beings to be ``interpreters" of sensory inputs by being attracted to rewards and repulsed by penalties, even though their first ancestors did not know what was beneficial or harmful in the surrounding environment.

Artificial General Intelligence (AGI) or strong AI has been pursued for many years by researchers in many fields, with the goal of reproducing human-level intelligence in machines, e.g., the ability of generalization and self-adaptation. So far, the AI scientific community has achieved outstanding results for specific tasks, i.e., weak or narrow AI. Such narrow AI implementations require highly specialized high performance computing systems for the training process. In this work, we propose to tackle the quest for general intelligence in its simplest form through evolution. It is therefore essential to develop a mutable environment that mimics what the first living beings with the simplest nervous systems faced. We define the simplest form of artificial general intelligence as the ability of an organism to continuously self-learn and adapt in a continuously changing environment of increasing complexity. Our definition of self-learning covers the meaning of self-supervised learning \cite{sermanet2018time} and self-reinforcement learning \cite{hilleli2018toward}. Therefore, a self-learning agent is capable of interpreting the reactions of the surrounding environment affected by the agent's actions, then the sensory inputs of the agent are utilized to supervise or reinforce itself. This occurs because the sensory inputs contain cues for the agent to learn on its own. For example, the pain after touching a candle's flame (as the authors experienced in their childhood).

In this work, we propose the Neuroevolution of Artificial General Intelligence (NAGI) framework. NAGI is a bio-inspired framework which uses plausible models of biological neurons, i.e., spiking neurons \cite{izhikevich2003simple}, in an evolved network structure that controls a sensory-motor system in a mutable environment. Evolution affects the connection structure of neurons, their neurotransmitters (excitatory and inhibitory), and their local bio-inspired learning algorithms. The inclusion of such learning algorithms under evolutionary control is an important factor to generate long-term associative memory neural networks which may have cells with different plasticity rules \cite{grewe2017neural}. Moreover, the genotype of the NAGI's agents does not contain the synaptic strength (weights) of the connections to avoid any innate knowledge about the environment. However, controllers that are selected for reproduction are those rewarded for their ability of self-learning and adaptation to new environments, i.e., an artificial newborn brain of an agent is an ``empty box" with the innate ability to learn how to survive in different environments during its lifetime. That is somewhat similar to how humans have a specialized brain part for learning quickly any language when they are born \cite{Zador582643}.

The remainder of this paper is organized as follows. It provides background knowledge for the proposed NAGI framework, and then describes related works. Afterward, it contains a detailed explanation of the conceptual framework. Finally, we conclude this work by discussing the relevance of our approach for current AGI research, and we elaborate on possible future works which may include such a novel bio-inspired parallel and distributed learning method.

\section{Background}
\label{sec:background}

The proposed NAGI framework brings together several key approaches in artificial intelligence, artificial life, and evolutionary robotics \cite{doncieux2015evolutionary}, briefly reviewed in this section. A Spiking Neural Network (SNN) is a type of artificial neural network that consists of biologically plausible neuron models \cite{izhikevich2003simple}. Such neurons communicate with spikes or binary values in time series. SNNs incorporate the concept of time by intrinsically modeling the membrane potential within each neuron. Neurons spike when the membrane potential reaches a certain threshold. When the signals propagate as neurotransmitters to the neighboring neurons, their membrane potentials are therefore affected, increasing or decreasing. While SNNs are able to learn through unsupervised methods, i.e., Hebbian learning \cite{hebb-organization-of-behavior-1949} and Spike-Timing-Dependent Plasticity (STDP) \cite{li2014activity}, spike trains are not differentiable and cannot be trained efficiently through gradient descent. NeuroEvolution of Augmenting Topologies (NEAT) \cite{stanley2002evolving} is a method that uses a Genetic Algorithm (GA) \cite{holland1992genetic} to grow the topology of a simple neural network and adjust the weights of the connections to optimize a target fitness function, while allowing to keep diversity (speciation) in the population and to maintain compatible gene crossover with historical marking. The neuroplasticity used to adapt the weights in the proposed NAGI framework includes the Hebbian learning rule as STDP. In particular, the weight adaptation of STDP happens when the neuron produces a spike or action potential through the axon (i.e., output connection). Such an event allows the modification of the synaptic strength of the dendrites (i.e., input connections) that caused or did not cause that spike.

Funes and Pollack \cite{funes1998evolutionary} describe the body/brain interaction (sensors and actuators vs. controller) as ``chicken and egg" problem; the course of natural evolution shows a history of body, nervous system, and environment all evolving simultaneously in cooperation with, and in response to, each other \cite{mautner2000evolving}. Embodied evolution \cite{Watson1999EmbodiedEE} is an evolutionary learning method for training agents through embodiment, i.e., embodied agents learning in an environment. Thus, in nature, general intelligence is a result of evolved self-learning through embodiment. 

\section{Related work}
\label{sec:related}

The idea of neuroevolution with adaptive synapses is not new. Stanley et al. \cite{stanley2003evolving} present NEAT with adaptive synapses using Hebbian local learning rules, with the goal of training neural networks for controlling agents in an environment. The authors verify the difference in performance with and without adaptation on the dangerous food foraging domain. The differences between their approach and ours are that their environment is static throughout the agent lifetime and they have inheritance of synaptic strength. Their results show that both networks with and without adaptive synapses reach the maximum fitness on that domain, and therefore both present ``adaptation". An extended version of the previous method is Adaptive Hypercube-based NEAT (HyperNEAT) \cite{risi2010indirectly}. Adaptive HyperNEAT includes indirect encoding of the network topology as large geometric patterns.

A recent work that uses NEAT and no weight inheritance is Weight Agnostic Neural Networks (WANN), introduced by Gaier and Ha \cite{gaier2019weight}. WANN is tested successfully with different supervision and reinforcement learning tasks whose weights are randomly initialized. Such promising results demonstrate that the network topology is as important as the connection weights in an artificial neural network. This is one of the motivations for the development of the NAGI framework. WANN and NEAT with adaptive synapses have been shown to be successful methods. However, such methods miss an important component for self-learning, which is a mutable environment as proposed in this work. 

A recent review of neuroevolution can be found in \cite{Stanley2019} and it shows how competitive NEAT and its extensions are in comparison to deep neural networks trained with gradient-based methods for reinforcement learning tasks. Neuroevolution provides several extensions, which include indirect encoding to allow scalability, novelty search to promote diversity, meta-learning for training a network to learn how to learn, and the combination with deep learning for searching deep neural network architectures. Furthermore, its authors envisage that neuroevolution will be a key factor to reach AGI through meta-learning and open-ended evolution. However, in NEAT the neural weights are inherited, so there is no explicit target for general intelligence and adaptation. 

A framework for the neuroevolution of SNNs and topology growth with genetic algorithms is proposed by Schaffer \cite{Schaffer2015}, with the goal of pattern generation and sequence detection. Eskandari et al. \cite{eskandari2016evolving} propose a similar framework for artificial creature control, where the evolutionary process modifies and inherits the network topology and the SNN weights to perform a given task.

A method which tries to produce general intelligence incrementally is PathNet \cite{fernando2017pathnet}, where deep neural network paths are selected through evolution to perform the forward propagation and weight adjustment. Such evolving selection allows the network to learn new tasks faster by re-using frozen (previously learned) paths without catastrophic forgetting. Another framework that tries to produce low-level general intelligence is described by Voss \cite{voss2007essentials}. It is a functional proof-of-concept prototype, owned by the company Adaptive A.I. Inc., which can interact with virtual and real world through sensors and actuators. Its controller, which has conceptual general intelligence capabilities, consists of a memory to save all data and to store the proprietary cognitive algorithms.

Multi-agent environments have also been considered a valuable stepping-stone towards AGI because the behavior of agents must adapt to cooperate and compete among them. One of the first examples of such multi-agent environment is the PolyWorld ecological simulator, introduced by Yaeger \cite{yaeger1994computational}. PolyWorld is a simulated environment of randomly generated food where evolving artificial organisms controlled by neural networks with Hebbian learning live. Organisms are able to eat, mate, fight, move, change the field of view, and utilize body brightness as a form of emergent communication. Their emergent behaviors are to some extent similar to the ones found in nature. Another recent multi-agent environment is presented by Lowe et al. \cite{lowe2017multi}. However, in such an environment, the adaptation occurs in the actor-critic methods of their reinforcement learning framework. Such method outperforms traditional reinforcement learning approaches on competitive and cooperative multi-agent environments. Another reinforcement learning method which exploits multi-agent environments is introduced by Jaderberg et al. \cite{jaderberg2018human}. In their work, they use the environment of Quake III Arena Capture the Flag, a 3D first-person multiplayer game. Their method in this game exceeds human-level performance, therefore the artificial agents are able to cooperate and compete among them and even with human players. One work that provides an open-source competitive multi-agent environment for research purposes is Neural MMO \cite{suarez2019neural}. Here, the agents are players which need to survive and prosper in an environment similar to the ones used in Massively Multiplayer Online Role-Playing Games (MMORPGs).

One important aspect of natural evolution is the ability to endlessly produce diverse solutions of increasing complexity, i.e., open-ended evolution (OOE). In contrast, OOE is difficult to achieve in artificial systems. A conceptual framework for the implementation of OOE in evolutionary systems is presented by Taylor \cite{taylor2019ooe}. Embodiment plays a key role in OOE in the context of the agent and its morphology, as discussed by Bongard \cite{bongardooe}. For an articulated summary and discussion of OOE see \cite{taylor2016ooe}. In \cite{stanley2017ooe} the authors argue that open-ended evolution is a grand challenge for artificial general intelligence, and artificial life methods are promising starting points for the inclusion of OOE in AI systems.

\section{Framework concept}
\label{sec:framework}

The main concept of the proposed NAGI framework is to mimic as close as possible the evolution of general intelligence in biological organisms, starting from the simplest form. To do that, we propose a minimalistic model with the following components. An agent is equipped with a randomly-initialized minimal spiking neural network. The agent is placed in a mutable environment in order to be able to generalize (learn to learn), instead of merely learning to solve the specific environment. Agents are more likely to survive if they perform correct actions. Agents have access to the environment through sensory inputs. The environment also provides intrinsic rewards and penalties. New agents inherit the topologies of the controllers from the previous generation (untrained), with the possibility of complexification (e.g., new neurons and synapses can appear through genetic operators). Training or learning happens throughout a generation. The goal of the untrained inherited controllers is to possess a topology that supports the ability to learn new environments. Neural learning occurs by utilizing environmental information (sensory input and environmental rewards/penalties) and neuroplasticity (e.g., Hebbian learning through spike-timing-dependent plasticity). The expected result is an unsupervised evolving system that learns without explicit training in a self-learning manner through embodiment. In this section, the components of the conceptual framework are described in details.

\subsection{Data representation}
\label{sec:data}

The data that flows to and from the spiking neural networks that control the agents are encoded as firing rate, i.e., the number of spikes per second. The firing rate has minimum and maximum values, and is represented for simplification as real number between 0 and 1 (i.e., range $[0.0,1.0]$). The stimulus to the neural networks can be Poisson-distributed spikes which have irregular interspike intervals, as observed in the human cortex \cite{heeger2000poisson}. That representation can be used for encoding input from binary environments (e.g., binary numbers $0$ and $1$ or Boolean values $False$ and $True$), or multi-value environments (e.g., represented as grayscale from black to white), and allows for representation of minimum and maximum activation values of sensors and actuators.

\subsection{Self-Learning through Embodiment}
\label{sec:embodied}

A new agent learns through the reactions of an environment via embodiment (i.e., by having a ``body" that affects an environment while sensing it). As such, the input of the neural network controller includes reward and penalty information for the learning process, such as the collision sensor in an autonomous robot whose activation represents a penalty, and a reward otherwise. This feedback information is the key factor for achieving self-learning. The concept of self-adaptation is closely connected with embodied cognition, a core property of living beings \cite{smith2005development}. In contrast, supervised learning and reinforcement learning use the error of the neural network output to globally adjust the network model through methods of iterative error reduction, such as gradient descent. In embodied learning, the input itself is used to adjust the agent's controller. Such sensory input contains the reactions of the environment to the actions of the agent. 

\begin{figure*}[!ht]
\includegraphics[width=\textwidth]{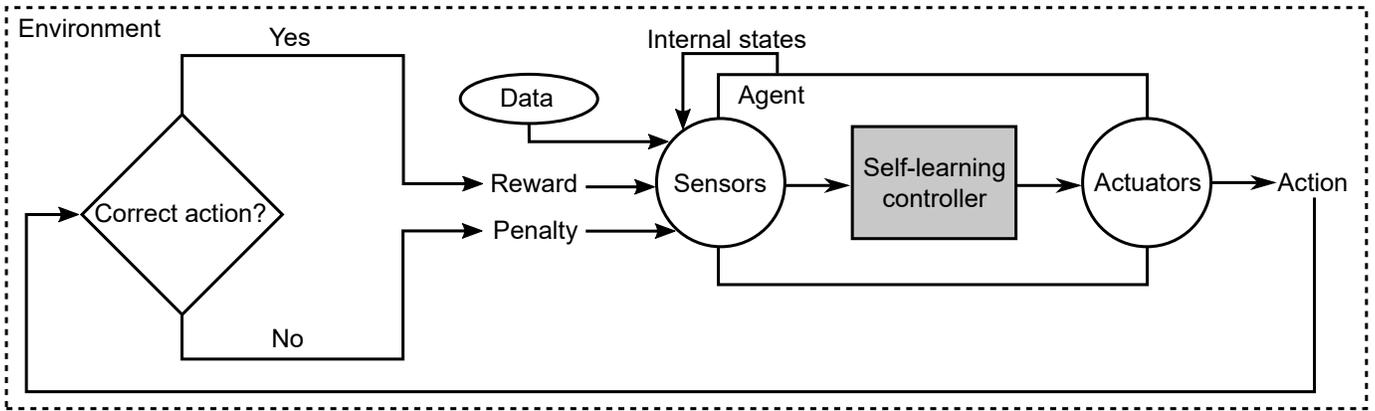}
\centering
\caption{Illustration of virtual embodied learning or self-learning through embodiment in a non-reactive environment. In the case of a reactive environment, rewards and penalties are embedded within the environmental data.}
\label{fig:embodied}
\end{figure*}

In the proposed framework, the local learning rules of the spiking neural network controller are responsible to correct the global behavior of the network according to agent experiences. This learning approach is, therefore, a result of self-learning through embodiment. The framework overview is depicted in Fig.~\ref{fig:embodied}. Note that self-learning through embodiment only works with agents in reactive environments (environments that affect the agents and are affected by them), such as any sensory-motor system deployed in the real-world. Non-reactive environments, on the other hand, do not react to any action of the agent, like any image classifier or object detector which only gives environmental information, thus there is no mutual interaction between an agent and a non-reactive environment. Therefore, we propose to create a virtual reactive environment for such cases. Virtual Embodied Learning (VEL) is the proposed method for such cases when no reward and penalty feedback is available through the sensory input. VEL adds reward and penalty inputs to a given sensory-motor system as illustrated in Fig.~\ref{fig:embodied}. It can also include the internal states of the agents, such as hunger and health. In addition, VEL can substitute supervised and reinforcement learning by using the loss of the model as penalty input and the opposite of the loss as reward input.

\subsection{Mutable environment}
\label{sec:environment}

To truly exploit and assess the self-learning capabilities and the generalization of the evolving spiking neural network, a mutable environment is proposed. The evolutionary goal of agents is to survive the changes in their environment. In the real-world, living organisms inherit modifications to their body and/or behavior through the generations. For example, a species may evolve a camouflage, such as the stick insects \cite{lev2004plant}, and another one may evolve the appearance of a poisonous or venomous animal, such as the false coral snakes \cite{davidson1996united}. The proposed mutable environment is a simple metaphor of such examples.

\begin{figure*}[ht]
\includegraphics[width=0.8\textwidth]{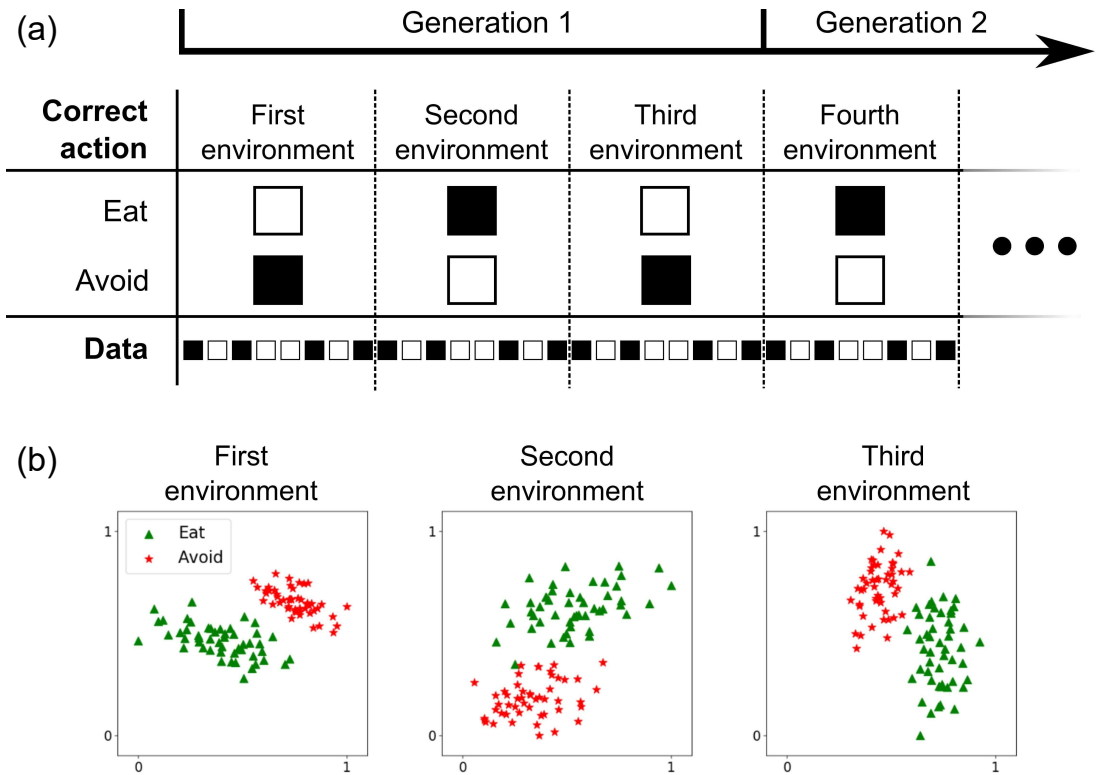}
\centering
\caption{Samples of mutable environments that can be presented to the agents through their evolution. Agents can execute two actions (eat or avoid). Within each generation, after the agents of a generation have seen all samples of an environment, a new one is presented. (a) 1D environment where the agent has one binary sensor. (b) The agent has two non-binary sensors (i.e, the axes).}
\label{fig:mutable_env}
\end{figure*}

Fig.~\ref{fig:mutable_env}a shows mutable environments that every agent in the population faces during its lifetime. Each agent has one sensor which provides one bit of information (i.e., $black$ or $white$) and can perform two actions (i.e., $eat$ or $avoid$). In each generation, the agents are presented with environmental data from several environments. Each sample is presented for a given period to allow the agents' controllers to learn. In the first environment, the correct action $eat$ is associated with the $white$ color while the action $avoid$ is associated with $black$. Once the environmental data has been consumed by the agent, there is an abrupt change in the interpretation of the environment ($black$ and $white$ are flipped) and the agents are presented with the environmental data again. Agents that perform well in many environments within each generation are more likely to go through the next generation.

Fig.~\ref{fig:mutable_env}b presents more complex mutable environments where agents have two sensors and non-binary environmental values can be received. As shown in the figure, different environments are procedurally generated and presented in each generation, where abrupt changes in the labeling of correct and wrong actions have happened. The set of actions may also be expanded to more than two, with different effects on agents' lifetime and their fitness scores.

\subsection{Neuroplasticity}
\label{sec:Neuroplasticity}

Each neuron in the evolved spiking neural network may have a different plasticity rule. The different types of learning rules are subject to evolutionary control. Examples of learning rules include asymmetric Hebbian, symmetric Hebbian, asymmetric anti-Hebbian, symmetric anti-Hebbian \cite{li2014activity}. Together with all the Hebbian learning rules encoded in the genome, there will be the effectiveness of the potentiation and the depression of the synapse strengths, i.e., how strong the learning rules are going to be for reducing or increasing the weight of the synapses. Moreover, other types of learning rules discovered in neuroscience may be added together or in parallel to those, such as non-Hebbian learning, neuromodulation, and synapse fatigue \cite{kato2009non,johansen2014hebbian,abrahamsson2005synaptic}. The neuroplasticity will also be regulated by a maximum total value of synaptic strength that a neuron can have for its dendrites. In case this value is reached, the increase in the weight of a synapse will cause a decrease of the others in the same neuron. This type of weight normalization is reported in \cite{royer2003conservation,el2018locally} for biological neurons.

\subsection{Neuroevolution}
\label{sec:Neuroevolution}

The population of genomes (spiking neural network controllers) for the agents is evolved through a modification of NEAT \cite{stanley2002evolving}. The genotypes of NEAT describe the topology and weights of the synapses, while our proposed method does not evolve the weights while includes in the genotype the type of neurotransmitter and neuroplasticity \cite{li2014activity}. The weights of the spiking neural networks are randomly initialized in every generation because the agents should not have innate knowledge of the environment \cite{Zador582643}. Therefore, the proposed framework focuses on the self-learning capabilities of the agents. Their lifetime will be longer when agents perform correct actions and shorter when they perform wrong actions. The lifetime of agents is used as fitness score to define the best performing neural networks.

Algorithm \ref{algo:evaluate} explains how an agent's genome is evaluated while it is in a mutable environment during its lifetime. The fitness score for the agent is equal to the time the agent is alive until its death. Each agent has a maximum life expectancy. Such life expectancy is reduced faster when an agent receives a penalty and it is reduced slower when the agent receives a reward. Both penalties and rewards reduce the lifetime of agents, as one agent is not to live for an infinite amount of time if it always performs the correct action.

The neuroevolution process allows the growth of neural network topologies and therefore the population is initialized with minimal networks that complexify over time. Nevertheless, there may be a penalty on lifetime to avoid the generation of big networks which may have neuron groups that specialize for each different environment. Therefore it allows the network to learn how to forget the previous environment, and then be able to adapt to the new one \cite{benjamin2011successful}. Another reason to apply this penalty for the size of the network is that more neurons require more energy to maintain them. This reduction of lifetime caused by the number of neurons can be regulated by a parameter, therefore choosing it is of high importance to the fitness and lifetime of the agent.

\begin{algorithm*}
\caption{Agent's genome evaluation using mutable environment and virtual embodied learning}
\label{algo:evaluate}
\begin{algorithmic}[1]
    \Procedure{evaluate}{$genome$}
    \State ${agent}\gets$ new Agent($genome$)
    \Comment{Agent is initialized with untrained neural network}
    \State $lifetime\gets 0$
    \While{$agent$ is alive}
        \If{$dataset$ is empty}
            \State $dataset\gets getNextDatasetForCurrentGeneration()$
            \Comment{Next temporary environment of the current generation}
        \EndIf
        \State $data,label\gets getRandomSampleAndDeleteFrom(dataset)$
        \State $reward\gets False$ 
        \Comment{Initialization of $reward$}
        \State $penalty\gets False$
        \Comment{Initialization of $penalty$}
        \While{$(agent$ is learningSample$)\land (agent$ is alive$)$}
        \Comment{Agent learns the presented sample with neuroplasticity for a period of time}
            \State $lifetime\gets lifetime+1$
            \State $action\gets agent(data,reward,penalty)$
            \State $reward\gets action=label$
            \State $penalty\gets action\ne label$
            \State $agent.healthReduction(reward,penalty)$
            \Comment{Penalty reduces the agent's health faster than reward, then accelerating its death}
        \EndWhile
    \EndWhile
    \State \Return $lifetime$
    \EndProcedure
\end{algorithmic}
\end{algorithm*}

\section{Discussion and Conclusion}
\label{sec:conclusion}

While current AI methods such as deep learning and reinforcement learning (and their combinations) have proven to be successful in solving a multitude of challenging tasks, e.g., defeating humans in the real-time strategy game Starcraft II \cite{alphastarblog}, there is a lot of debate around the limitation of current methods for breakthroughs in Artificial General Intelligence. One key difference between AI and AGI is the learning ability. Most of AI methods (supervised, unsupervised, and reinforcement learning) are explicitly trained, while AGI needs some intrinsic ability to self-learn. 

One of the open questions for AGI research is: how can artificial agents be able to acquire the general skill of learning, in order to continuously adapt throughout their lifetime? 

In biological systems, we infer that self-learning is a result of rewards and penalties which are embedded in the sensory data living beings receive from the environment (unlabeled and mutable data). Their ability to learn through this form of self-learning through embodiment is a result of evolution.

One of the goals of the proposed NAGI framework is an AGI system that allows the adaptation and general learning skills through the three main levels of self-organization in living systems \cite{sipper1997phylogenetic}:

\begin{itemize}
  \item Phylogeny, which includes evolution of genetic representations;
  \item Ontogeny, which takes care of the morphogenetic process (growth) from a single cell to a multicellular machine, by following the genotype instructions;
  \item Epigenesis, which allows the emergence of a learning system through an indirect encoding between genotype and phenotype, and the phenotype is subject to modifications (learning) throughout the lifetime while interacting with the environment.
\end{itemize}

We, therefore, envision the proposed spiking neural network model will include developmental and morphogenetic processes \cite{doursat2012morphogenetic} in future extensions of the framework.

Another envisioned stepping-stone to AGI is the extension of the framework to artificial life multi-agent systems. Multi-agent environments will allow the emergence of more advanced strategies of adaptation and learning based on collaboration and competition. 
In addition, the framework may benefit from extending the environment itself into an evolving agent, which can also allow for increased complexity and open-ended evolution.

Finally, we expect that future implementations of the NAGI framework and its extensions will be deployed/embodied into real robot agents equipped with physical sensors.

In conclusion, this work proposes a novel general framework for the neuroevolution of artificial general intelligence (NAGI) in its simplest form, which can be extended to more complex tasks and environments. In NAGI, the general intelligence, i.e., learning to learn to adapt to different environments, is a result of self-learning through embodiment. Therefore, the learning process is not a result of explicit training with supervision or reinforcement learning, as there is no loss function used to adjust the neural network weights. The proposed neural network model is a bio-inspired model based on spiking neural networks. Their learning is based on spike-timing-dependent plasticity which uses only input data for learning. As such, penalties and rewards are embedded within the environmental data sensed by the agents.  

This work describes the details of the NAGI conceptual framework as a novel paradigm for self-learning. Therefore, the experimental results are not included in this contribution. However, our current experimental results are promising, and part of a separate contribution. 

The NAGI conceptual framework proposes a computational system which may allow the simplest form of general intelligence observed in nature to emerge. Self-learning through embodiment shifts the way machines currently learn by changing the paradigm of supervised and reinforcement learning. Our efforts are also to reduce the gap between biological neural networks computation and artificial intelligence implementations, allowing for a biologically-inspired neural network model that suits the paradigm in artificial life of massively parallel, distributed, and local interactions. 

\section*{Acknowledgements}
We thank Gustavo Moreno e Mello, Kristine Heiney, Anis Yazidi, and Benedikt Vogler for thoughtful comments and discussions. This work was supported by Norwegian Research Council SOCRATES project (grant number 270961).

\bibliographystyle{IEEEtran}
\bibliography{IEEEabrv,bib}

\end{document}